\documentclass[runningheads]{llncs}
\usepackage{graphicx}
\usepackage{color}
\usepackage{subfig}
\usepackage{graphicx}
\usepackage{amsmath,amssymb,amsfonts}
\usepackage{algorithmic}
\usepackage{textcomp}
\usepackage{textgreek}
\usepackage{braket,upgreek,bbold}
\usepackage{mathtools}

\begin{document}
\title{Comparing Machine Learning based Segmentation Models on Jet Fire Radiation Zones}
\titlerunning{Comparing ML based Segmentation Models on Jet Fire Radiation Zones}
%
\author{Carmina Pérez-Guerrero\inst{1} \and
Adriana Palacios\inst{2} \and
Gilberto Ochoa-Ruiz\inst{1} \and
Christian Mata\inst{3} \and
Miguel Gonzalez-Mendoza\inst{1} \and
Luis Eduardo Falcón-Morales\inst{1}
}
\authorrunning{C. Pérez-Guerrero et al.}
%
\institute{Tecnológico de Monterrey, School of Engineering and Sciences, Mexico. \and
Universidad de las Americas Puebla, Department of Chemical, Food and Environmental Engineering, Puebla, 72810, Mexico.
\and
Universitat Politècnica de Catalunya. EEBE, Eduard Maristany 16, 08019 Barcelona. Catalonia, Spain.
}
\maketitle         
\begin{abstract}
Risk assessment is relevant in any workplace, however, there is a degree of unpredictability when dealing with flammable or hazardous materials so that detection of fire accidents by itself may not be enough. An example of this is the impingement of jet fires, where the heat fluxes of the flame could reach nearby equipment and dramatically increase the probability of a domino effect with catastrophic results. Because of this, the characterization of such fire accidents is important from a risk management point of view. One such characterization would be the segmentation of different radiation zones within the flame, so this paper presents exploratory research regarding several traditional computer vision and Deep Learning segmentation approaches to solving this specific problem. A data set of propane jet fires is used to train and evaluate the different approaches and given the difference in the distribution of the zones and background of the images, different loss functions, that seek to alleviate data imbalance, are also explored. Additionally, different metrics are correlated to a manual ranking performed by experts to make an evaluation that closely resembles the expert’s criteria. The Hausdorff Distance and Adjusted Rand Index were the metrics with the highest correlation and the best results were obtained from the UNet architecture with a Weighted Cross-Entropy Loss. These results can be used in future research to extract more geometric information from the segmentation masks or could even be implemented on other types of fire accidents.

\end{abstract}
\section{Introduction}
\label{intro}

In the industrial environment, there are certain activities such as the storage of fuels or the transportation of hazardous material, that can be involved in severe accidents that affect the industrial plant or activity border, as well as external factors like human health, environmental damage, property damage, and others. Overall knowledge of the features and characteristics of major accidents is required to prevent and manage them or, in the worst case scenario, take action to reduce and control their severity and aftermath. Some accidents are relatively well known and researched, however, other accidents have not been broadly explored.

Sometimes the detection of fire, in an industrial setting, is not enough to make the correct decisions when managing certain fire accidents. An example of this, is the impingement of jet flames on nearby pipes, depending on the heat fluxes of the flame, the pipe wall could heat up quickly and reach dangerous temperatures that lead to severe domino effect sequences. These heat fluxes can be defined as three main areas of interest within the flame \cite{Vahid21}, and the definition of their geometric characteristics and localization becomes valuable information in risk management.

\begin{figure}[!htbp]
\centering
\subfloat[]{\includegraphics[width = 1.8in]{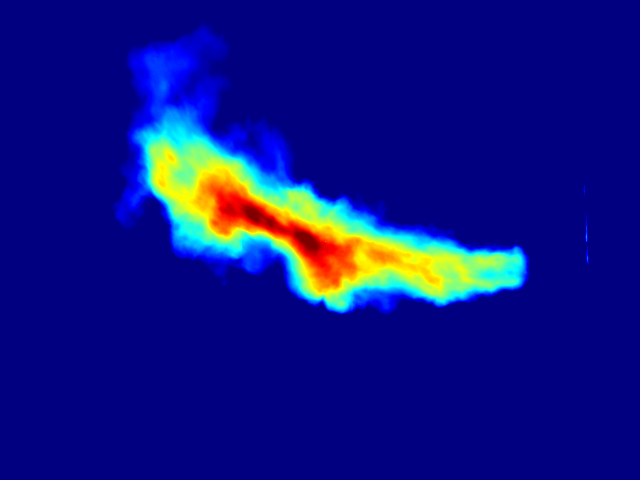}}
\qquad
\subfloat[]{\includegraphics[width = 1.8in]{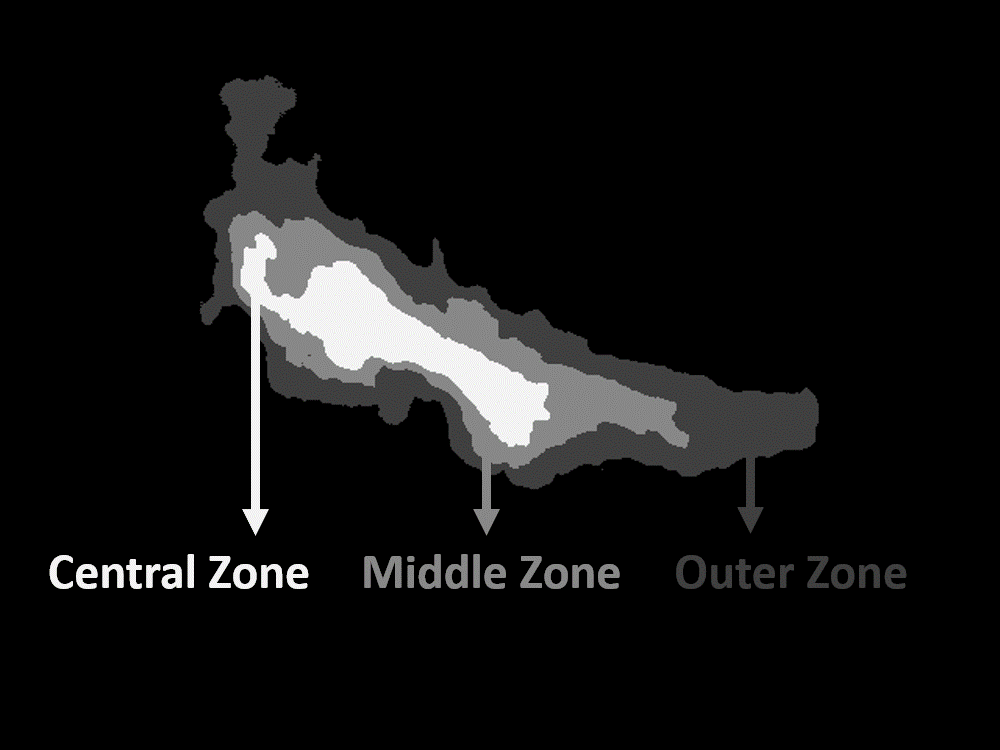}}
\caption{Image (a) is an infrared visualization of a horizontal propane jet flame. Image (b) is the corresponding ground truth segmentation, with the segment names indicated. Modified from \cite{Vahid21}.}
\label{fig:horizontal}
\end{figure}

Semantic segmentation could be a useful approach for this kind of flame characterization, and the same procedure could even be employed in other fire-related accidents. To explore this proposal, different segmentation methodologies are evaluated on a set of images from real jet fires of propane to accurately segment the radiation zones within the flames. These different zones are illustrated in Figure \ref{fig:horizontal} and are defined across this paper as the Central Zone, the Middle Zone, and the Outer Zone.

The rest of this paper is organized as follows. Section \ref{sota} describes the previous work present in the literature related to the problem and different semantic segmentation methods. Section \ref{approach} describes the approaches used to perform the exploration research. Section \ref{dataset} explains the data set used for the experiments and the pre-processing methods applied. Section \ref{sec:metrics} describes the different evaluation metrics and loss functions explored during the experiments. Section \ref{training} contains the training protocol used for the experiments with Deep Learning architectures. Section \ref{testing} explains the testing procedure applied to the segmentation methods. Section \ref{results} presents the results of the exploration research. This section also offers a discussion of the future work that can stem from the knowledge obtained. Finally, Section \ref{conclusions} summarizes the major findings of the presented work.
 
\section{State of the Art}
\label{sota}

There has been previous research work regarding the use of Computer Vision and image processing for the detection and monitoring of flares in the context of industrial security. For example, Rodrigues and Yan \cite{Rodrigues11} used imaging sensors combined with digital image processing to determine the size, shape, and region of the flare, successfully characterizing the dynamism of the fire. 

Another example is the work done in Janssen and Sepasian \cite{Janssen18}, where, by separating the flare from the background using temperature thresholding, and adding false colors to represent different temperature regions, a system was created, capable of tracking the flare size for automated event signaling.

\subsection{Deep Learning Architectures}
\label{sec:architectures}

Deep learning algorithms, such as Convolutional Neural Networks, have shown outstanding performance in many complex tasks, such as image recognition, object detection, and semantic and instance segmentation \cite{Litjens_2017}. Advancements on these methods have been increasing rapidly, with constant research being published regarding better and more robust algorithms. Some important architectures present in the literature are summarized in Table \ref{tab:architecture-sum} with their advantages and disadvantages described.  In general, these architectures were selected due to their efficiency during segmentation and their capacity to accurately portray the shape of such dynamic figures obtained from the flames.

\begin{table}[!htbp]
\caption{Summary of the selected architectures.}
\begin{tabular}{llll}
\hline
\textbf{Source      } & \textbf{Architecture      } & \textbf{Pros}  & \textbf{Cons}\\ \hline
\cite{chen17}            & DeepLabv3             & \begin{tabular}[c]{@{}l@{}}Recovers detailed structures \\ lost due to spatial invariance.\\ Efficient approximations via \\ probabilistic inference.\\ Wider receptive fields.\end{tabular} & \begin{tabular}[c]{@{}l@{}}Low accuracy on small \\ scaled objects.\\ Has trouble capturing \\ delicate boundaries of objects.\end{tabular}                                                                                             \\ \hline
\cite{badrinarayanan16}            & SegNet                & \begin{tabular}[c]{@{}l@{}}Efficient inference in terms of memory \\ and computational time.\\ Small number of trainable parameters.\\ Improved boundary delineation.\end{tabular}          & \begin{tabular}[c]{@{}l@{}}Input image must be fixed.\\ Performance drops with a \\ large number of classes and  \\ is sensitive to class imbalance.\end{tabular}                                                                                                        \\ \hline
\cite{ronneberger15}            & UNet                  & \begin{tabular}[c]{@{}l@{}}Can handle inputs of arbitrary sizes.\\ Smaller model weight size.\\ Precise localization of regions.\end{tabular}                                                                      & \begin{tabular}[c]{@{}l@{}}Size of network comparable \\ to the size of features.\\ Significant amount of time \\ to train.\\ High GPU memory footprint \\ in larger images.\end{tabular} \\ \hline
\cite{oktay18}            & \begin{tabular}[c]{@{}l@{}}Attention \\ UNet\end{tabular} & \begin{tabular}[c]{@{}l@{}}Avoids the use of multiple similar \\ feature maps.\\ Focuses on the most informative features \\ without additional supervision.\\ Enhances the results of the UNet architecture. \end{tabular}                                                                                                                             & \begin{tabular}[c]{@{}l@{}}Adds more weight parameters \\ The time for training increases, \\ especially for long sequences\\ \end{tabular} \\ \hline

\end{tabular}
\label{tab:architecture-sum}
\end{table}

\subsection{Traditional Computer Vision Methods}
\label{sec:traditional}

Different traditional segmentation methods are included in the analysis as a baseline for the results given by the Deep Learning architectures, these methods are Gaussian Mixture Model (GMM), K-means clustering, Thresholding, and Chan-Vese segmentation.

\begin{itemize}
    \item GMM is one of the main methods applied to fire and smoke image segmentation. It offers a clear definition of their dynamic shape, but sometimes missed pixels from the inner parts of the fire and smoke \cite{Ajith19}.
    \item K-means clustering provides shape-based image segmentation \cite{Zaitoun15} and has been previously used for fire segmentation \cite{Rudz13} \cite{Ajith19}. However, this method does not guarantee continuous areas \cite{Zaitoun15}.
    \item Thresholding is the simplest method of image segmentation and has a fast operation speed. It is most commonly used in region-based segmentation, but it is sensitive to noise and gray scale unevenness \cite{yuheng17}.
    \item The Chan-Vese segmentation \cite{Chan01} belongs to the group of active contour models, which present some advantages for infrared image segmentation, since the edges obtained are smooth and are represented with closed curves. However, this kind of segmentation is very sensitive to noise and depends heavily on the location of the initial contour, so it needs to be manually placed near the image of interest \cite{Zhang16}.
\end{itemize}

\section{Proposed Approach}
\label{approach}

To explore the semantic segmentation of radiation zones within the flames as characterization of fire incidents, a group of 4 traditional segmentation methods and 4 deep learning architectures are explored. These segmentation methods are trained and tested using a data set of jet fire images obtained from videos of an experiment performed in an open field.

To properly evaluate and compare the different segmentation approaches, several metrics with different evaluation methods are correlated to manual rankings performed by two experts in the field, this is to make sure that the evaluation is the most representative of a fire engineer's perception of good segmentation. 

The best model found from this exploratory research will then be used in future work, not included in this paper, to extract other geometric characteristics from the resulting segmentation masks.

\section{Data set}
\label{dataset}

Investigators from Universitat Politècnica de Catalunya performed an experiment to produce horizontal jet fires at subsonic and sonic gas exit rates. The experiment was filmed using an infrared thermographic camera, more specifically, an AGEMA 570 by Flir Systems. The video was saved in four frames per second, resulting in a total of 201 images with a resolution of 640 x 480 pixels. After obtaining the infrared visualizations of the flames, segmentation of the three radiation zones within the fire was performed, the results were validated by experts in the field and the result was a ground truth for each of the infrared images. The images were saved as Matlab files that contain a temperature matrix corresponding to the temperature values detected by the camera for the infrared images, and the label matrix corresponding to the segments for the ground truth segmentations. These files were then exported as PNG files to be used by the different segmentation algorithms described in Section \ref{sec:architectures}. 
\subsection{Image Processing}
To enhance the characteristics of the jet fires represented in the infrared images, and to reduce their variance, a process of image normalization was employed, which also helps in the convergence of the Deep Learning methods. The ground truth images were transformed into labeled images, where a label id was used instead of the original RGB values.
Given the small number of samples of the data set, and to increase the variability of the input during the training of the models, data augmentation techniques were applied. This processing of the images can increase the performance of the models, can avoid the over-fitting to the training samples, and can help the models to also perform well for instances that may not be present in the original data set. Horizontal flipping, random cropping, and random scaling were applied in parallel to the training workflow, so for each iteration of training, a different augmented image was inputted. The probability of horizontal flipping was set to 50\%, and random scaling had values that ranged between 0.7 and 2.0.

\section{Metrics and Loss Functions}
\label{sec:metrics}

\subsection{Metrics}
An analysis was performed to select evaluation metrics that were more representative of an expert's evaluation of the segmentation. The metrics analysed are separated into groups that describe their evaluation method. This diversity is important because the consideration of particular properties could prevent the discovery of other particular errors or lead to over or underestimating them. To compare the metrics enlisted in Table \ref{tab:metrics}, a group of images were evaluated with the metrics and two manual rankings performed by two experts in the field. A Pearson pairwise correlation is used to perform this comparison at segmentation level.

\begin{table}[!htbp]
\centering
\caption{\label{tab:metrics} Summary of the metrics analysed in this paper. The "Group" column describes the method group that the metric belongs to. Based on \cite{taha15}.}
\begin{tabular}{ll}
\centering
\textbf{Metric}            & \textbf{Group}              \\ \hline
Jaccard Index              & Spatial Overlap Based       \\ \hline
F-measure                  & Spatial Overlap Based      \\ \hline
Adjusted Rand Index      & Pair Counting Based         \\ \hline
Mutual Information         & Information Theoretic Based \\ \hline
Cohen's Kappa              & Probabilistic Based         \\ \hline
Hausdorff Distance         & Spatial Distance Based      \\ \hline
Mean Absolute Error        & Performance Based           \\ \hline
Mean Square Error          & Performance Based           \\ \hline
Peak Signal to Noise Ratio & Performance Based           \\ \hline
\end{tabular}
\end{table}

\subsection{Loss Functions}
\label{sec:lossfunc}
The proportion of the different radiation zones within each flame is different. The Outer Zone tends to be the largest segment and the central zone is usually smaller than the Middle Zone, these differences could affect the overall segmentation obtained during training, which is why the loss functions employed in this research work were focused on dealing with this class imbalance. The implemented loss functions are summarized in Table \ref{tab:losses}.

\begin{table}[!htbp]
\centering
\caption{\label{tab:losses} Summary of the loss functions implemented in this paper.}
\begin{tabular}{lll}
\hline
\textbf{Source   } & \textbf{Loss Function}                                                              & \textbf{Description}                                                                                                                                                  \\ \hline
\cite{pytorch}           & \begin{tabular}[c]{@{}l@{}}Weighted Cross-Entropy\\ Loss\end{tabular}               & \begin{tabular}[c]{@{}l@{}}Combines Log Softmax and Negative Log Likelihood.\\ Useful for multiple classes that are unbalanced.\end{tabular}               \\ \hline
\cite{lin20}           & Focal Loss                                                                          & \begin{tabular}[c]{@{}l@{}}Addresses foreground-background class imbalance.\\ Training is focused on a sparse set of hard examples.\end{tabular}                      \\ \hline
\cite{fidon17}            & \begin{tabular}[c]{@{}l@{}}Generalized Wasserstein \\ Dice Loss (GWDL)\end{tabular} & \begin{tabular}[c]{@{}l@{}}Semantically-informed generalization of the Dice score.\\ Based on the Wasserstein distance on the probabilistic \\ label space.\end{tabular} \\ \hline
\end{tabular}
\end{table}

\section{Training}
\label{training}

\begin{figure}[!htbp]
\centering
\subfloat[Loss values for the DeepLabv3 model. Early Stopping took place at epoch 701.]{\includegraphics[width = 1in]{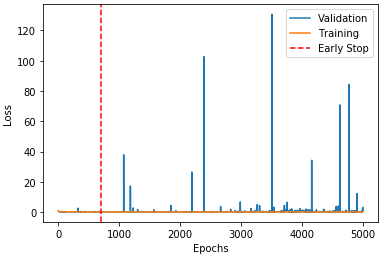}}
\qquad
\subfloat[Loss values for the SegNet model. Early Stopping took place at epoch 1274.]{\includegraphics[width = 1in]{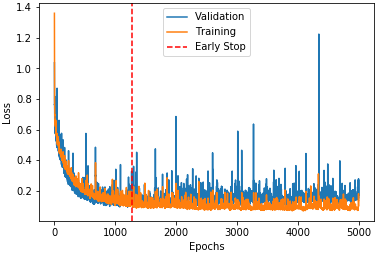}}
\qquad
\subfloat[Loss values for the UNet model. Early Stopping took place at epoch 1460.]{\includegraphics[width = 1in]{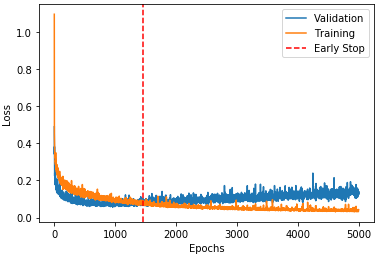}}
\qquad
\subfloat[Loss values for the Attention UNet model. Early Stopping took place at epoch 1560.]{\includegraphics[width = 1in]{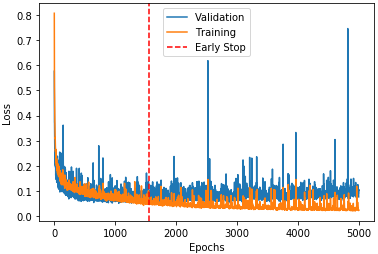}}
\caption{The Weighted Cross-Entropy loss values for the best models obtained for each architecture explored.}
\label{fig:losses}
\end{figure}

The PyTorch framework \cite{pytorch} was used for the implementation of the Deep Learning architectures. To maintain the weights of the Convolutional Neural Networks as small as possible, a weight decay strategy was used with L2 regularization. The learning rate had an initial value of 0.0001 and used an ADAM optimizer during training. The class weights used by the Weighted Cross-Entropy and Focal losses were computed according to the ENet custom class weighing scheme \cite{paszke16} and are defined as the following: 1.59 for background, 10.61 for Outer zone, 17.13 for Middle zone, and 22.25 for Central zone. The class distances used for the loss function of GWDL is defined as 1 between the background and the radiation zones, and as 0.5 between the zones themselves. The training was performed using an Nvidia DGX workstation that has 8 Nvidia GPUs and allows a batch size of 4 as maximum, which is the batch size used for the models. The data was split into 80\% for training and 20\%  for testing and validation, resulting in a total of 161 images for training, 20 images for testing, and 20 images for validation. The models were trained for up to 5000 epochs with an Early Stopping strategy to avoid overfitting.
The resulting loss values for the best models can be visualized in Fig. \ref{fig:losses}.

\section{Testing}
\label{testing}

Testing and Validation were performed on 20\% of the data set, which represent a total of 40 images. The results are compared using the metrics with the highest correlation to the manual ranking of experts. The results for each Deep Learning architecture are first compared across all 3 different loss functions mentioned in Section \ref{sec:lossfunc}. The best performing combination, for each architecture, is then compared to the results obtained from the other 4 traditional segmentation models. The time each method takes to segment the whole data set of 201 images is also taken into account. The goal of the comparison is to find the best overall model for the segmentation of the radiation zones within the flames.

\section{Results and Discussion}
\label{results}

\subsection{Selected Metrics}

\begin{figure}[!htbp]
  \includegraphics[scale=0.6]{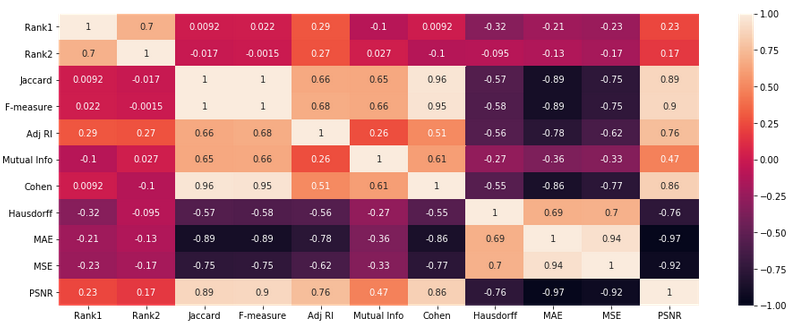}
  \caption{Heatmap representing the Pearson Correlation Coefficient for the metrics in Section \ref{sec:metrics} and the manual rankings Rank1 and Rank2 done by experts in the field of the problem.}
  \label{fig:corr}
\end{figure}

The results of the correlation between the metrics, mentioned in Section \ref{sec:metrics}, and the manual rankings performed by experts can be observed in Fig. \ref{fig:corr}. 

The highest Pearson correlation value was given by the Hausdorff Distance, with a value of -0.32 to the first manual ranking, this negative relationship is because a smaller Hausdorff Distance is preferable and the manual ranking assigned higher values as the segmentation improved. The second highest correlations were given by the Adjusted Rand Index in both manual rankings, with a value of 0.29 for the first one and 0.27 for the second one, this positive relationship is because the values of the Adjusted Rand Index go from 0 to 1, with 1 being the best result, and showing similar behavior to the manual ranking that assigns higher values to better segmentation.

\subsection{Best Loss Function}

\begin{figure}[!htbp]
\centering
\subfloat[Hausdorff Distance for all the DeepLabv3 models.]{\includegraphics[width = 1in]{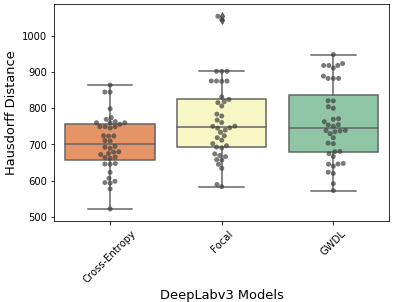}}
\qquad
\subfloat[Hausdorff Distance for all the SegNet models.]{\includegraphics[width = 1in]{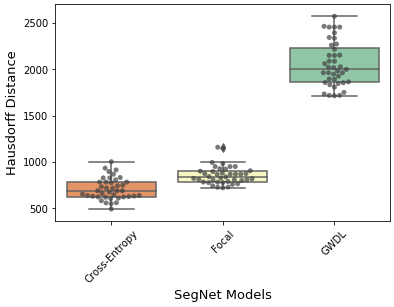}}
\qquad
\subfloat[Hausdorff Distance for all the UNet models.]{\includegraphics[width = 1in]{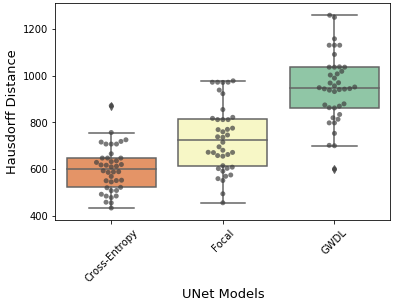}}
\qquad
\subfloat[Hausdorff Distance for all the Attention UNet models.]{\includegraphics[width = 1in]{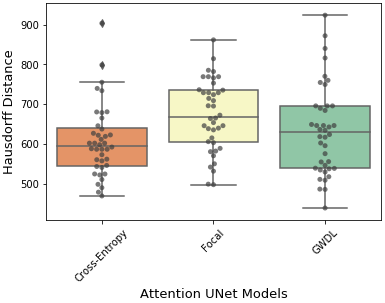}}
\caption{The Hausdorff Distance for all the model and loss function combinations across the validation set. For each model the order of loss functions is Weighted Cross-Entropy, Focal and GWDL from left to right.}
\label{fig:combinations}
\end{figure}

Each combination of Deep Learning architecture and Loss Function was evaluated using the Hausdorff Distance as the main metric. The best results across all architectures were obtained when using the Weighted Cross-Entropy Loss, this can be observed in Fig \ref{fig:combinations}. In general, the Focal Loss also showed good results, being close to the Weighted Cross-Entropy results and most of the times surpassing the GWDL results, with the only exception being the Attention UNet model, where the mean Hausdorff Distance with Focal Loss is larger than with GWDL, but has a much closer distribution. The most dramatic difference happened with the SegNet architecture, which showed very poor results with GWDL.

\subsection{Traditional and Deep Learning Segmentation}

\begin{figure}[!htbp]
\centering
\subfloat[Hausdorff distance for all segmentation models.]{\includegraphics[width = 2in]{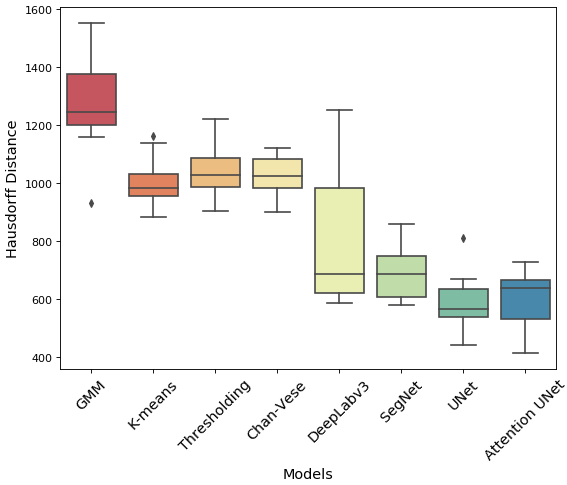}}
\qquad
\subfloat[Adjusted Rand Index for all segmentation models.]{\includegraphics[width = 2in]{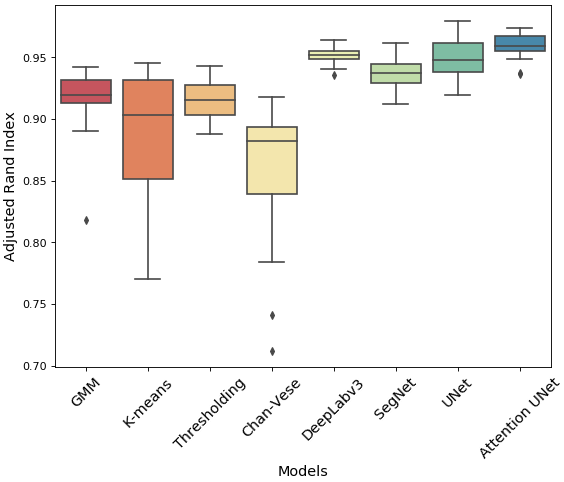}}
\caption{The Hausdorff Distance and Adjusted Rand Index values for all the segmentation models across the testing set.}
\label{fig:segmodels}
\end{figure}

A comparison was done between the traditional Computer Vision methods mentioned in Section \ref{sec:traditional} and the Deep Learning models mentioned in Section \ref{sec:architectures},using the Weighted Cross-Entropy loss. The Hausdorff Distance and the Adjusted Rand Index are used to evaluate the models across the testing set and the time each method takes to perform the segmentation on all the images from the data set is also taken into account. These results are summarized in Table \ref{tab:summary} and the distribution of the results can be visualized in Fig \ref{fig:segmodels}. Overall the best performing models are observed to be the UNet and Attention UNet models. Even if Attention UNet achieved a slightly better Adjusted Rand Index score, UNet obtained a better Hausdorff distance and segmentation time, therefore we can say that the best model is the one that uses the UNet architecture.

\begin{figure}[!htbp]
\centering
\subfloat[GMM.]{\includegraphics[width = 1in]{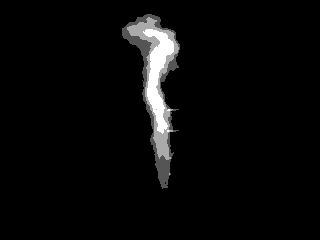}}
\qquad
\subfloat[K-means.]{\includegraphics[width = 1in]{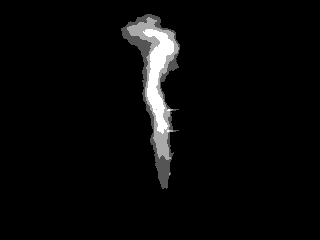}}
\qquad
\subfloat[Thresholding.]{\includegraphics[width = 1in]{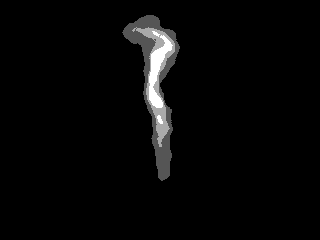}}
\qquad
\subfloat[Chan-Vese.]{\includegraphics[width = 1in]{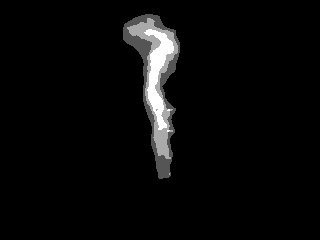}} \\
\subfloat[DeepLabv3.]{\includegraphics[width = 1in]{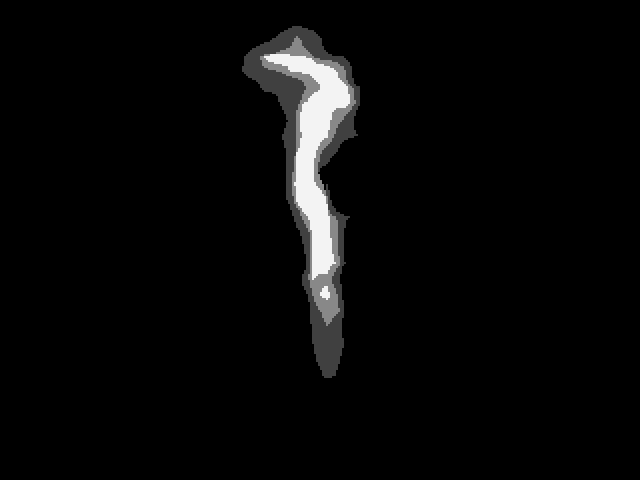}}
\qquad
\subfloat[SegNet.]{\includegraphics[width = 1in]{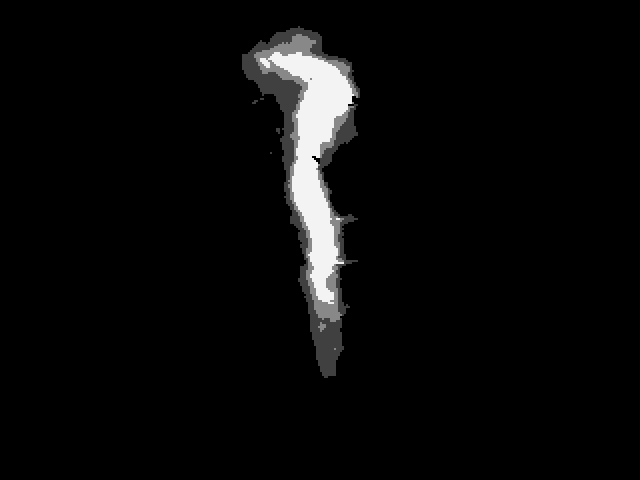}}
\qquad
\subfloat[UNet.]{\includegraphics[width = 1in]{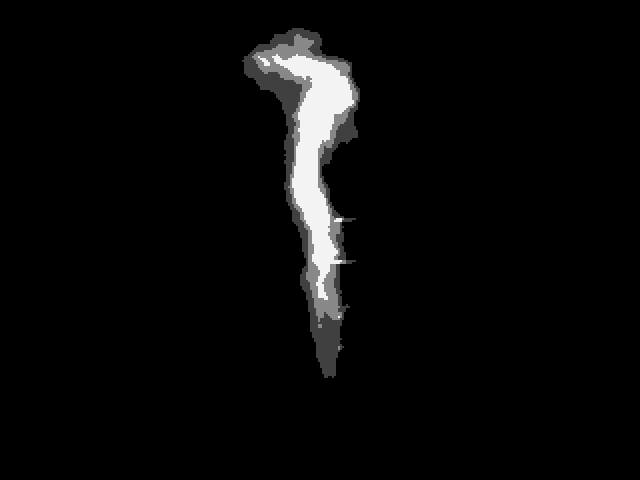}}
\qquad
\subfloat[Attention UNet.]{\includegraphics[width = 1in]{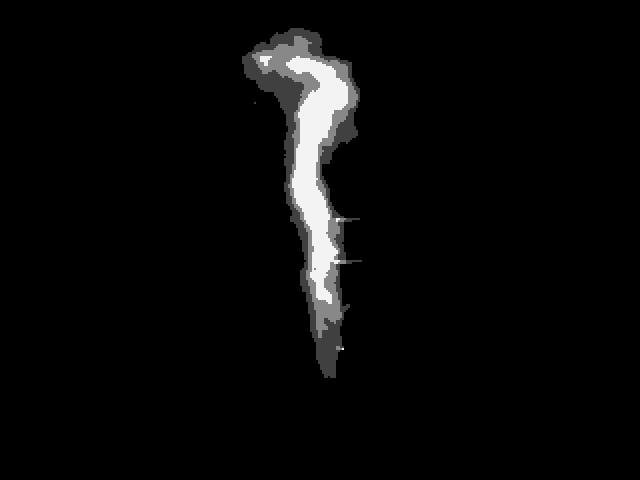}}
\caption{Sample segmentations of all the models.}
\label{fig:segmentations}
\end{figure}

The difference in the segmentation of each method can also be visualized in Fig \ref{fig:segmentations}. It can be observed that the shape of the Outer and Middle zones are generally well represented in most of the segmentation models, however, the most important differences are found in the Central zone, where the Deep Learning architectures defined more clearly its shape. Similar results of the UNet and Attention UNet architectures are also observed in this sample segmentation masks, with close resemblance to the ground truth.

\begin{table}[!htbp]
\centering
\caption{Mean Hausdorff Distance and Adjusted Rand Index values for all the segmentation models across the testing set, as well as the time in seconds that each method takes to segment the whole data set. The best results are in bold.}
\begin{tabular}{llll}
\hline
\textbf{Method} & \textbf{Hausdorff Distance    } & \textbf{Adjusted Rand Index   } & \textbf{Time (s)} \\ \hline
GMM             & 1288.10                     & 0.9156                 & 2723.8           \\ \hline
K-means         & 1000.63                     & 0.8855                 & 3035.1            \\ \hline
Thresholding    & 1029.08                     & 0.9152                 & 30.7            \\ \hline
Chan-Vese       & 1031.90                     & 0.8568                 & 18177.5              \\ \hline
DeepLabv3       & 784.86                      & 0.9514                 & 17.1              \\ \hline
SegNet          & 692.73                      & 0.9381                 & 16.4              \\ \hline
UNet            & \textbf{586.46}                      & 0.9504                 & \textbf{15.7}              \\ \hline
Attention UNet  & 601.05                      & \textbf{0.9592}                 & 17.7              \\ \hline
\end{tabular}
\label{tab:summary}
\end{table}

\subsection{Discussion}
Overall, the Deep Leaning algorithms greatly outperformed the traditional Computer Vision methods and the best-proposed model would be a UNet architecture with a Weighted Cross-Entropy loss function. The segmentation masks obtained from this model could be used in the future to further extract other geometric characteristics of the flame, such as length, area, and lift-off distance. This additional information would improve greatly the decision making process involved in the risk assessment and management of fire related accidents that can take place in an industrial setting. Furthermore, the metrics of Hausdorff Distance and Adjusted Rand Index can be used to evaluate other segmentation approaches that may try to solve similar problems in the future, having the certainty that the evaluation would be a close representation of an expert's opinion.

\section{Conclusions}
\label{conclusions}

The semantic segmentation of radiation zones within the flames can be used to characterize fire accidents, such as jet fires, and the information obtained from the segmentation can prove to be critical when dealing with risk management in industrial settings. The exploratory research presented in this paper continued to show that Deep Learning architectures greatly outperform other traditional Computer Vision approaches. It was also found that for this specific problem, the best loss function to train a Deep Learning model is a Weighted Cross-Entropy Loss, the best architecture to be used is UNet, and the best evaluation metrics are both the Hausdorff Distance and the Adjusted Rand Index. All this knowledge can be later used in future research focused on extracting even more geometric information from the segmentation masks. This could bring about a more complete characterization analysis of jet fires and the methods applied to these types of fire accidents could then be used on other fire scenarios.

\subsubsection*{Acknowledgements}
This research is supported in part by the Mexican National Council of Science and Technology (CONACYT). This research is part of the project 7817-2019 funded by the Jalisco State Council of Science and Technology (COECYTJAL). The data set that supports the findings of this study are available upon reasonable request.
%
%
%
\bibliographystyle{splncs04}
\bibliography{mybibliography}

\end{document}